\documentclass[conference]{IEEEtran}
\IEEEoverridecommandlockouts
\usepackage{cite}
\usepackage{amsmath,amssymb,amsfonts}
\usepackage{algorithmic}
\usepackage{graphicx}
\usepackage{textcomp}
\usepackage{xcolor}
\usepackage{url}
\usepackage[skip=1ex,labelfont=bf, textfont=it]{caption}
\usepackage[textfont=rm]{subcaption}

\def\BibTeX{{\rm B\kern-.05em{\sc i\kern-.025em b}\kern-.08em
    T\kern-.1667em\lower.7ex\hbox{E}\kern-.125emX}}
\begin{document}

\title{QLAMMP: A Q-Learning Agent for Optimizing Fees on Automated Market Making Protocols\\
}

\author{\IEEEauthorblockN{Dev Churiwala}
\IEEEauthorblockA{\textit{Viterbi School of Engineering*}\thanks{*Visiting Student from BITS Pilani, Goa Campus.}  \\
\textit{University of Southern California}\\
Los Angeles, USA \\
f20180602@goa.bits-pilani.ac.in}
\and
\IEEEauthorblockN{Bhaskar Krishnamachari}
\IEEEauthorblockA{\textit{Viterbi School of Engineering} \\
\textit{University of Southern California}\\
Los Angeles, USA \\
bkrishna@usc.edu}
}

\maketitle

\begin{abstract}
	Automated Market Makers (AMMs) have cemented themselves as an integral part of the decentralized finance (DeFi) space. AMMs are a type of exchange that allows users to trade assets without the need for a centralized exchange. They form the foundation for numerous decentralized exchanges (DEXs), which help facilitate the quick and efficient exchange of on-chain tokens. All present-day popular DEXs are static protocols, with fixed parameters controlling the fee and the curvature - they suffer from invariance and cannot adapt to quickly changing market conditions. This characteristic may cause traders to stay away during high slippage conditions brought about by intractable market movements. We propose an RL framework to optimize the fees collected on an AMM protocol. In particular, we develop a Q-Learning Agent for Market Making Protocols (QLAMMP) that learns the optimal fee rates and leverage coefficients for a given AMM protocol and maximizes the expected fee collected under a range of different market conditions. We show that QLAMMP is consistently able to outperform its static counterparts under all the simulated test conditions.
\end{abstract}


\section{Introduction}
With the ever-growing interest in and adoption of blockchain and cryptocurrency technology among common actors, institutions, and academicians, Decentralised Finance (DeFi) applications have rapidly gained popularity. DeFi’s rise preserves the general principles of scalability, security, and decentralization. One of its core advancements was the adoption of Decentralized Exchanges (DEXs) built atop the Automated Market Maker (AMM) model. This revolutionized DeFi, bringing efficient trading with no restrictions, complete anonymity, and quick settlement to the table. Academicians have recently begun exploring these DEXs and released Systematization of Knowledge works detailing the protocols' functioning~\cite{wang2020automated, xu2021sok, mohan2022automated}. We concern ourselves with these DEXs, which have also independently accounted for a staggering \$1.5 trillion in aggregated transaction volumes over the past year~\cite{DEXVolumes}.

Traditionally, most AMM protocols have had static parameters controlling their fee rate and curvature - together controlling the protocol's bid/ask prices, which indirectly govern the swap volume and, in turn, the total accrued fees. Let us consider the direct implications of such a model. Let's say the protocol is optimized to perform best when users swap large amounts with higher tolerances to loss. Now, consider the user demands changing (perhaps due to external market factors) in terms of their traded volume or slippage appetite, the protocol retains its previous invariant configuration and likely performs poorly. This paper aims to solve this dynamic user problem by adapting the fee rate and curve behavior based on market conditions. For this purpose, we propose a Reinforcement Learning (RL) agent that modifies the protocol parameters at regular intervals, referred to as Q-learning Agent for Market Making Protocols (QLAMMP).

RL is a well-established technique that has the ability to learn and build knowledge about dynamic environments~\cite{sutton1992reinforcement}. We propose the use of the Q-learning RL algorithm to train our agent~\cite{watkins1993technical}. By interacting with the environment, the proposed agent builds a knowledge base of the environment, actions, and expected rewards given the constantly changing conditions. In this way, the agent can converge on an optimal policy that maximizes the expected total rewards, in this case, the fees accrued by the protocol. Hence, Q-learning is a promising approach to learning policies in an uncontrolled environment, such as the ever-shifting demands of a DEX's user base. To our knowledge, this is the first time RL has been used to optimize fees on an AMM protocol.

The main contributions of this paper are as follows:
\begin{itemize}
    \item We are the first to study the use of an RL agent to optimize the fees on an AMM protocol by modifying fee rates and the curve behavior.
    \item We hypothesize and evaluate the agent's performance under a slew of different conditions. 
    \item We empirically evaluate the effect that the size of the updating interval (measured as an update made every $k$ number of steps) has on the agent's performance. 
    \item We also look at the standard deviation of the agent's actions through the training epochs to ensure that the agent is adapting to the environment and taking different actions rather than just optimizing for and sticking to a condition-agnostic state.
    \item To interpret QLAMMP's performance, we also break down the agent's behavior into action space components - a subset of the actions modifying just the fee rate, another subset of actions modifying just the curvature - thus seeing the effect that a particular subset of the actions has on the agent's performance.
    \item To accommodate for the dynamic user conditions and achieve robust agent performance under varied conditions - we develop a modular training environment, which can be adapted to fit numerous different AMM protocols - thus allowing for the agent's training based on specific use cases.
    \item We also touch upon the practical aspect of maintaining decentralization and privacy using a decentralized oracle or on-chain deployment of QLAMMP. 
\end{itemize}

The remainder of the paper is structured as follows. In Section II, we review the related work. Section III provides a basic background on AMMs needed to parse the rest of the paper. Section IV presents our proposed system model, and Section V describes the associated problem formulation. Section VI explains the simulation methodology used, and Section VII presents our results and findings. Finally, we conclude the paper in Section VIII.

\section{Related Works}
AMMs were popularized by Hanson's Logarithmic Market Scoring Rule (LMSR) for prediction markets~\cite{hanson2007logarithmic}. Since AMMs are well suited to providing continuous liquidity even in low market action scenarios, they are now widely employed as exchange curves in the financial trading domain. Bonding Curve AMMs, such as the ones we explore in our work, were first introduced by the Bancor Protocol. This AMM facilitated trades by pairing each token against their native protocol token, the (Bancor Network Token) BNT, with each non-native token having an independent liquidity pool with BNT~\cite{hertzog2017bancor}. 

Building on these projects, Buterin proposed a curve-based AMM for a DEX~\cite{buterin_2017}, which was later implemented by Adams \emph{et al.} in the form of Uniswap, which has cemented itself as one of the biggest DEXs in the cryptocurrency space~\cite{adams_zinsmeister_robinson_2020}. Subsequently, Egorov proposed and developed the Curve Finance Protocol, another major DEX that focuses on efficient trades for tokens with equivalent pricing~\cite{egorov2019stableswap}.  

The popularity of these DEXs led to much interest in the academic community. Angeris and Chitra analyzed curve-based AMMs to model, formalized the mathematical properties of these market makers, and formulated optimal arbitrage by traders as a convex optimization problem~\cite{angeris2020improved}. This was followed by the work of Evans \emph{et al.}, who developed a framework to optimally select fees for Geometric Mean Market Makers (G3Ms) when a modeled continuous process governed the price dynamics~\cite{evans2021optimal}.

Wang, Xu \emph{et al.} and Mohan have also released works on the 
Systemization of Knowledge (SoK) for DEXs using AMMs. These works provide a comprehensive view of the recent advances in the DEX space, an intuition for how different curves function, and also go on to explain the differences and similarities in the various underlying mathematical models~\cite{wang2020automated, mohan2022automated,xu2021sok}.

Krishnamachari \emph{et al.} have recently formulated a novel solution to eliminate arbitrage opportunities in curve-based AMMs. They propose a construction of AMMs using \textit{dynamic curves}. The proposed curve uses an external market oracle to match the pool pricing to the market pricing continuously - they show that doing so eliminates arbitrage opportunities while retaining greater liquidity in the token pool~\cite{krishnamachari2021dynamic}. Wang and Krishnamachari built on this, formulating a method to find an optimal trading policy for dynamic curve AMMs using a dynamic programming approach~\cite{wang2022optimal}.

RL is a sub-domain of Machine Learning (ML) first introduced by Sutton \emph{et al.} to solve problems formulated as Markov decision processes (MDPs), a formulation of decision-making problems~\cite{sutton1992reinforcement}. RL has been used by several researchers in decision-making problems pertaining to blockchain systems. Qiu \emph{et al.} train an agent to maximize the offloading performance of computationally expensive tasks (mining/data processing) in blockchains~\cite{qiu2019online}. Liu \emph{et al.} propose an RL framework to maximize transactional throughput for blockchain-enabled Internet of vehicles (IoV) data~\cite{liu2019deep}. In Sattarov \emph{et al.}, they develop strategies to trade cryptocurrencies using historical price data and current market prices~\cite{sattarov2020recommending}. Hou \emph{et al.} proposed an RL framework to analyze attacks on blockchain incentive mechanisms, such as selfish mining and block withholding attacks~\cite{hou2019squirrl}. Al-Marridi \emph{et al.} train an RL agent to solve for a Healthcare system's blockchain-configuration mapping with the goal of maximizing security while minimizing delay and cost~\cite{al2021smart}. Mani \emph{et al.} used RL to optimize a market maker on the traditional stock markets. They optimize for profit, inventory, and market quality~\cite{mani2019applications}. However, none of the past works explore an RL framework for optimizing fees in market-making protocols. In this paper, we propose such an RL framework to optimize the fees on AMM protocols. In particular, we train a Q-learning agent to adjust the protocol's parameters dynamically to maximize the fees collected over time.

\section{Background on AMMs}
Recent advances in developing decentralized exchanges (DEXs) have primarily used Automated Market Makers (AMMs) at their cores~\cite{adams_zinsmeister_robinson_2020, egorov2019stableswap, adams2021uniswap}. First used as price aggregation agents for prediction markets, AMMs act as an independent counterparty with pools of different tokens. Users looking to trade tokens can do so by directly buying and selling from the AMM protocol, which generates buy \& sell prices algorithmically. Thus, AMMs do away with the traditional order book system of market making and can provide liquidity even when market action is limited.

\subsection{Fundamental Properties}
\subsubsection{Swap}
A swap refers to an exchange of tokens between a user and a counterparty; here, the AMM protocol. In the context of a DEX, it involves the user querying the DEX to check the bid/ask spread, the slippage, and the loss to fees and then deciding whether or not they want to make the trade with the protocol.
\subsubsection{Bonding Curve}
Each AMM depends on its underlying bonding curve to generate trading prices algorithmically. This bonding curve is defined by a conservation function that follows an invariant property. To understand this, consider the following function.
\begin{equation}
    xy = k
\end{equation}
Here, constant $k$'s value is determined by the initial liquidity provision (when the system is initialized); using this $k$ and the new quantity of a token, the reserve quantity of the other token can be solved - thus giving us a buy/sell price.

\subsubsection{Slippage}
Slippage, for AMMs, is defined as the percent difference between the spot price and the realized price of a trade. This difference comes about due to two main factors, (1) the change in the bid/ask spread during trade execution and (2) the relative size of the swap compared to the total liquidity in the system. The slippage for a given protocol is particularly pronounced when the liquidity in the system is low or when a swap amount represents a significant fraction of the liquidity pool. 

\subsection{Exchange Functions}
\subsubsection{Constant Sum Market Marker}
Although not widely used due to its glaring deficiencies, the most basic elementary type of exchange function is the Constant Sum Market Maker (CSMM). For token reserves $(x, y)$, a CSMM maintains the sum of reserves constant; that is, the exchange function satisfies the following equation.
\begin{equation}
    (R_x - \delta_x) + (R_y + \delta_y) = k
\end{equation} 
Of course, one could generalize this to multiple tokens by maintaining the following equation. 
\begin{equation}
    \sum_{i=1}^{n} R_i = k
\end{equation}
\subsubsection{Constant Product Market Maker}
Due to its robustness under different market conditions, many DEX protocols currently use a Constant Product Market Maker (CPMM) bonding curve. For token reserves $(x, y)$, a CPMM maintains the product of reserves constant; that is, the exchange function satisfies the following equation.
\begin{equation}
    (R_x - \delta_x) * (R_y + \delta_y) = k
\end{equation}
Of course, one could generalize this to multiple tokens by maintaining the following equation. 
\begin{equation}
    \prod_{i=1}^{n} R_i = k
\end{equation}
\subsubsection{Hybrid Function Market Maker}
The Hybrid Function Market Maker (HFMM) combines the best of CSMMs and CPMMs. It draws on the fact that there is no price slippage when using a CSMM and that the protocol can provide liquidity even when the markets are limited, like a CPMM. The HFMM does this by combining the exchange functions of both CSMMs and CPMMs. 

Therefore, given the same initial conditions, the bonding curve for a HFMM lies somewhere between a CSMM and the CPMM curves based on how it is set up. The curvature of an HFMM can be controlled based on a leverage coefficient $\mathcal{A}$ - which decides the skewness of the curve toward either the CPMM curve or the CSMM curve.

\subsection{Noteworthy Protocols}
\begin{itemize}
    \item \textit{Uniswap}: The now popular, Uniswap protocol, introduced in 2018, had only one pair of tokens per smart contract, and both tokens had the same value. Under these assumptions, the protocol maintained a constant product invariant to facilitate exchanges on the Ethereum Blockchain~\cite{adams_zinsmeister_robinson_2020}.
    \item \textit{Curve Finance}: The Curve Finance protocol, formerly called StableSwap, had liquidity pools consisting of two or more assets with the same peg, for example, USDC and DAI. This approach allows Curve to use more efficient algorithms and feature the lowest levels of fees, slippage, and impermanent loss of any DEX on the Ethereum Blockchain~\cite{egorov2019stableswap}.
\end{itemize}

\begin{figure*}[ht]
\centering
    \setkeys{Gin}{width=\linewidth}
\centerline{\includegraphics{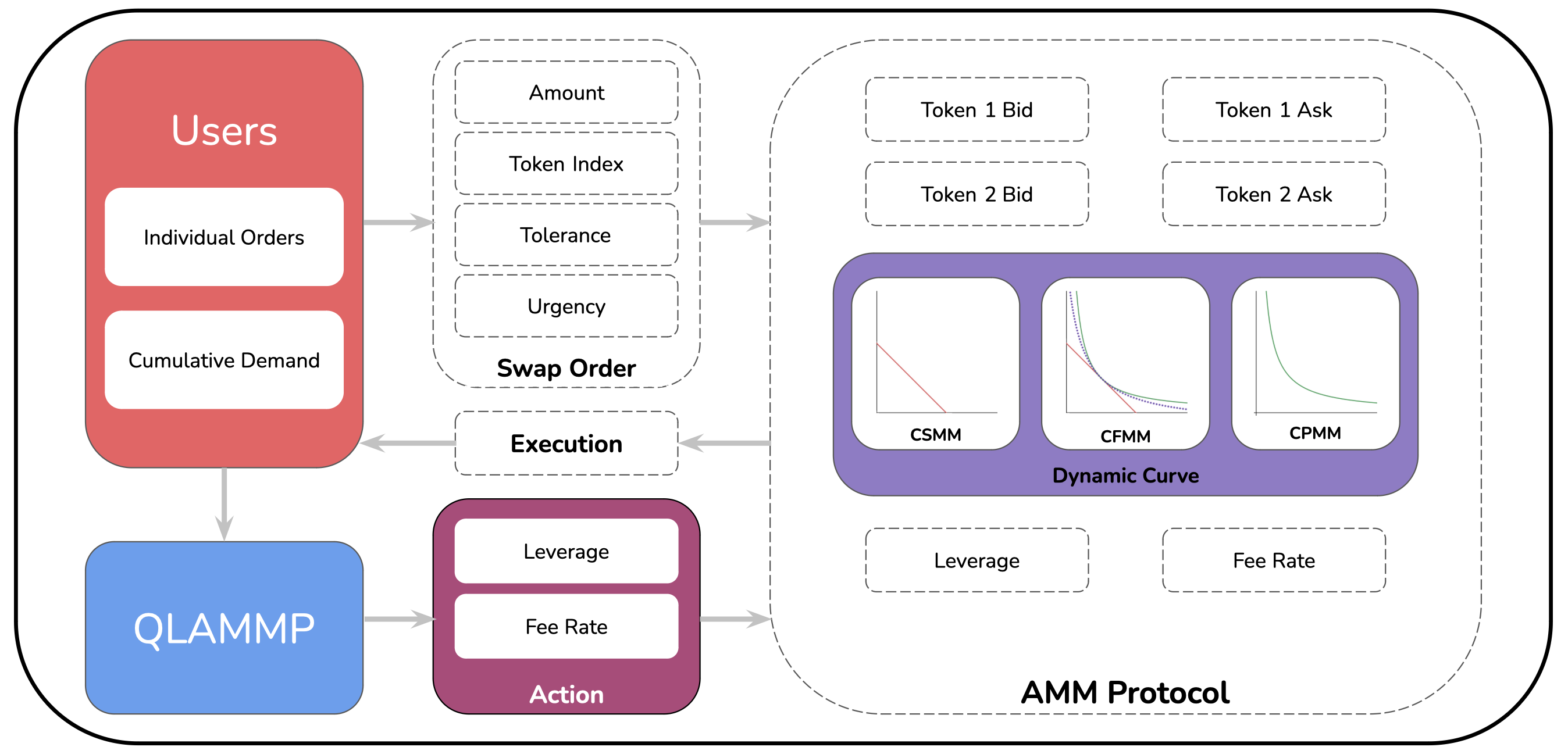}}
\caption{System model visualization}
\label{figimg}
\end{figure*}

\section{System Model}
In this section, we first provide an overview of the system and discuss the hybrid automated market-making protocol used for our simulations. Next, we present the proposed user dynamics and, finally, the swap dynamics - how a user on the protocol carries out each swap.

\subsection{Overview}
Our system consists of a central AMM protocol that acts as a counter-party to all swaps. The system also has users who request swaps. The users' decision to swap is based on modeled demand based on the value they get for their tokens. These users query the protocol to check the bid/ask price and the expected gain/loss on the desired swap. The swaps are modeled as a queue, with the front being serviced first. The AMM protocol aims to maximize its collected fee within the system.

\subsection{Protocol Dynamics}
We design our proposed Q-Learning Agent to optimize for a hybrid automated market-making protocol. This protocol is inspired mainly by the whitepaper for Curve Finance, primarily designed to facilitate stablecoin trades. The protocol's underlying exchange function can dynamically shift from a constant-product invariant to a constant-sum invariant. This shift is accomplished by interpolating between the two invariants, controlled by a Leverage Coefficient, \(\mathcal{A}\). When \(\mathcal{A}\) → 0, the exchange function emulates a constant-product one (like Uniswap); when \(\mathcal{A}\) → +$\infty$, the exchange function is essentially a constant-sum one. This behavior is captured by the following invariant proposed in~\cite{egorov2019stableswap}:

\begin{equation}
 \mathcal{A}n^{n}\sum x_i + D = \mathcal{A}Dn^{n} + \frac{D^{n+1}}{n^{n}\prod x_i}
\end{equation}

Our software implementation of the protocol is based on the implementation by the recent survey on AMMs~\cite{xu2021sok}. We modify their implementation by adding fee mechanics to the Curve class. For each successful swap, the protocol collects a set percent of fee, $\mathcal{R}_f$, from the input tokens, $\mathcal{T}_i$, to get the final tokens, $\mathcal{T}_f$.

\begin{equation}
\mathcal{F} = \mathcal{T}_i * \frac{\mathcal{R}_f}{100}
\end{equation}
\begin{equation}
\mathcal{T}_f = \mathcal{T}_i - \mathcal{F}
\end{equation}

We define a \texttt{getLoss}() function, allowing users to query the protocol for the current loss, or price imapact, $\varphi$, resulting from slippage and fees. Additionally, we define new accessor methods to make it easier for objects of the user class to interact with the curve protocol.

\begin{equation}
\varphi = \frac{\mathcal{T}_f - \Delta \sum x_i}{\mathcal{T}_f} * 100
\end{equation}

\subsection{User Dynamics}
We have designed our simulation to closely emulate a real user interacting with a financial exchange, making swapping decisions based on empirical factors such as price and slippage and a subjective component - urgency. This modeling is random for each swap, with each parameter of the swap being sampled from carefully chosen random distributions. Assuming there are 
$\mathcal{N}$ users in the system, we let $U = \{U_1, U_2, ..., U_\mathcal{N}\}$, denote the set of users. Each user, $i$ $(i \in U)$, has a non-negative balance of the two tokens, $\{x_k, y_k\}$ - initialized to 1,000 each, and a number identifying them, $k$.

The environment generates swaps (mechanics discussed in Section~\ref{userGen}) and assigns them to random users. Upon receiving such a swap order, the user randomly chooses a token index, either 0 or 1. If the balance of the chosen index is less than a given fraction (set to 20\% by default) of the other token, indices are inverted. This mechanism prevents the drying up of either token’s balance. Once the swap parameters are finalized, the user queries the protocol for the expected $\varphi$. The swap goes through if the returned $\varphi$ is less than the urgency-modified tolerance. If not, the swap is canceled based on a 40\% chance modified by the urgency factor. If the swap is urgent, it is less likely to get canceled and vice-versa. In addition, if the swap is not canceled, the urgency is increased by 1\%, and the “\textit{holding}” swap status is returned to the environment.

\subsection{Swap Dynamics}
A swap object, $\mathcal{E}_i$, as described by us, consists of a slippage tolerance value, $s_i$, an urgency value, $u_i$, and a user identification number, $k$ $(k \in \{1, 2, ...\ \mathcal{N}\})$. The environment generates a set number of swaps based on the number of users in the system. The environment samples normal distributions for each swap’s tolerance and urgency value. Each of these swaps is randomly assigned to users in the system and pushed into a queue, ready to be executed. If the swap goes through or is canceled by the user, it is popped from the queue.

These swap objects also hold an amount, token index, and a new flag for swaps that were returned with a “\textit{holding}” status by the user - the object is then pushed back $k$ places (by default $k$ = 10) places in the swap queue and goes through again after the protocol’s state has changed over the ten transactions. If the swap is tried a predetermined number of times (by default set to 15) times and fails with a “\textit{holding}” status, it is automatically canceled by the environment and popped from the queue.

The swap method finally returns either,

    (1) \texttt{status code 1} → amount; ``\textit{success}",
    
    (2) \texttt{status code 0} → 0; ``\textit{holding}",
    
    (3) \texttt{status code -1} → -1; ``\textit{canceled}".

\section{Problem Formulation}
We frame the problem in a way such that one can infer cause and effect relationships from the simulations. Consequently, we propose three different agents:

(1) Agent 1, whose actions only change the fees for the protocol.

(2) Agent 2, whose actions can only modify the leverage coefficient for the protocol.

(3) QLAMMP, whose actions can modify the fees and the leverage coefficient.

These agents allow us to observe the effects of each control vector individually and the combined effect of both. 

\subsection{State Space}
At each decision epoch, $t$, the state space for the system can be characterized by $\mathcal{S}_t = (U,\ \mathcal{E}_t,\ \mathcal{L}_t,\ \mathcal{A},\ \mathcal{R}_f)$. The meaning of each variable is as follows.
\begin{itemize}
    \item $U = \{U_1, U_2, ..., U_\mathcal{N}\}$, denotes the set of users in the system, each user, $i$ $(i \in U)$, has their current non-negative balance of the two tokens, $x_i, y_i$.
    \item $\mathcal{E}_t = current\ swap$ denotes the swap at the head of the swap queue, it consists of a slippage tolerance value, $s_i$, an urgency value, $u_i$, and a user identification number, $k$ $(k \in \{1, 2, ...\ \mathcal{N}\})$.
    \item $\mathcal{L}_t = slippage$ denotes the protocol predicted slippage for the given swap, $\mathcal{E}_t$. Since this is a continuous - real value, we discretize it.
    \item $\mathcal{A} = leverage\ coefficient$ that controls the bonding curve's curvature.
    \item $\mathcal{R}_f = fee\ rate$ denotes the amount of fee collected by the protocol on each swap. 
\end{itemize}
This state space is shared across the aforementioned agents. Upon getting a swap instruction from the environment, the user generates a token index and an amount, which are used to query the protocol for the expected $\varphi$. $\varphi$ consists of two components, (1) the current fee collected, $\mathcal{F}$, and (2) the current slippage. As the agent chooses actions to adapt the fee rate, $\mathcal{R}_f$, and the leverage coefficient, $\mathcal{A}$ - it attempts to optimize $\varphi$ to allow for maximum fee collection and minimum trade cancellations.

\subsection{Action Space}
\subsubsection{Agent 1}
This agent has control over the rate of fees of the protocol. This fee rate is doubly bound with a lower limit of 0.04\% and an upper limit of 0.30\%. These values are derived from the real-world rates of Curve Finance and Uniswap V2, the largest exchanges of their kind, respectively. This range ensures representative taxation on each transaction. 

At each decision epoch, $t$, the agent can either increase the fee rate by 1, decrease the fee rate by 1, or leave it as it was previously. This change is governed by the limits mentioned above.

\subsubsection{Agent 2}
This agent controls the protocol's leverage coefficient, $\mathcal{A}$. This coefficient is doubly bound with a lower limit of 0 and an upper limit of 85. Once again, these limits are derived from the real-world coefficients of Uniswap V2 ($\mathcal{A} = 0$) and Curve Finance ($\mathcal{A} = 85$). This range ensures representative exchange dynamics over all the transactions. 

At each decision epoch, $t$, the agent can either increase $\mathcal{A}$ by 2, decrease $\mathcal{A}$ by 2, or leave it as it was previously. This change is governed by the limits mentioned above.

\subsubsection{QLAMMP}
This agent controls the protocol's leverage coefficient, $\mathcal{A}$, as well as its rate of fees. These parameters are bound by the same limits as Agents 1 \& 2. In this way, we can get representative simulations - which can also be compared against each other.  

At each decision epoch, $t$, the agent can make one of nine actions. These are a cross-product of the three actions of Agent 1 and the three actions of Agent 2. The larger action domain gives QLAMMP broader control over the protocol and potentially a better shot at optimizing for fees. Once again, all these changes are governed by the limits mentioned above. 

\subsection{Rewards}
The reward mechanism is shared across the three agents. We model the system by defining three different reward types that are awarded under different conditions.
\subsubsection{Swap Success} 
At a decision epoch, $t$, the user, $U_i$, specified by swap, $\mathcal{E}_t$, is able to make the swap with the expected slippage, $\mathcal{L}_t$, is less than $U_i$'s slippage tolerance, $s_i$. 

The agent is awarded a positive value equal to the fee procured by the swap.
    
\subsubsection{Swap Holding} 
At a decision epoch, $t$, the user, $U_i$, was pushed back into the swap queue, $\mathcal{E}$, due to the expected slippage, $\mathcal{L}_t$, being beyond $U_i$'s slippage tolerance, $s_i$.

The agent is awarded 0 because the outcome of this swap remains to be decided.
 
\subsubsection{Swap Canceled} At a decision epoch, $t$, the user, $U_i$, cancels the swap, $\mathcal{E}_t$, either due to random chance or because its number of total queries for the swap crossed 15. 

In such a case, the agent is awarded -1. 

These three rewards cases ensure that the agent is rewarded for enabling swaps, and maximizing the procured fee. At the same time, the agent is highly penalized for swaps being canceled. This harsh penalty is because, a canceled swap is a direct loss of fees that could have potentially been earned.

\section{Simulation Methodology}
Our simulation environment is written in Python using the OpenAI Gym toolkit~\cite{van1995python, brockman2016openai}. We opt for an object-oriented methodology to allow for modularity amongst the agents and the environment. The program consists of the protocol classes, the user class, the swap class, and the main environment class.

\subsection{Reinforcement Learning Algorithm}
We utilize the Q-Learning algorithm to train the agent on the given problem. It is a model-free off-policy RL method where the agent's goal is to obtain an optimal action-value function $Q_*(s, a)$ by interacting with the environment. It maintains a state-action table $Q[S, A]$ called a Q-table containing Q-values for every state-action pair. At the start, Q-values are initialized to all zeroes. Q-learning updates the Q-values using the Temporal Difference method~\cite{watkins1993technical}.

\begin{equation}
    Q(s_t, a_t) \xleftarrow{} Q(s_t, a_t) + \alpha(R_{t+1} + \gamma\max_aQ(s_{t+1}, a_t) - Q(s_t, a_t))
\end{equation}

where $\alpha$ is the learning rate and $\gamma\in[0,1]$. $Q(s_t, a_t)$ is the actual Q-value for state-action pair $(s_t, a_t)$. The target Q-value for state-action pair $(s_t, a_t)$ is \{$R_{t+1} + \gamma\max_aQ(s_{t+1}, a_t)$\} i.e. immediate reward plus discounted Q-value of next state. The table converges using iterative updates for each state-action pair. To efficiently converge the Q-table, an $\epsilon$-greedy approach is used.

The $\epsilon$-greedy approach: At the start of the training, all Q-values are initialized to zero. This uniformity implies that all actions for a state have the same chance to be selected. So to enable the convergence of the Q-table using iterative updates, the exploration-exploitation trade-off is employed. The explorative update for the Q-value of random state-action pair $(s_t, a_t)$ is carried out by randomly selecting the action. The exploitative update selects a greedy action $(a_t)$, with maximum expected rewards, for the state $(s_t)$ from the Q-table.

So, to converge the Q-table from the initial condition (all Q-values are set to zero), the agent initially prioritises exploration and later it prioritises exploitation. It uses probability $\epsilon$, to choose random action, or it chooses an action from the Q-table using probability $1 - \epsilon$. In the beginning, the value of $\epsilon$ is one, and it decays with time, tending to zero as the Q-table converges.

\begin{equation}
    \epsilon = \epsilon_{min} + (\epsilon_{max} - \epsilon_{min}) * e^{-\eta * \chi}
\end{equation}

where $\epsilon_{max}\ \&\ \epsilon_{min}$ are the maximum and minimum values for $\epsilon$, predefined by the environment; and $\eta$ is the factor by which $\epsilon$ is reduced every epoch, $\chi$.

\subsection{Simulation Setup}
We consider a dynamic exchange function as the underlying curve to facilitate trades on our Automated Market Maker (AMM). The AMM is initialized with 20,000 units of the two tokens. We populate the system with 20 users and 400 swaps per epoch. Under normal swap conditions, each user is given 1,000 units of each token at the beginning of the epoch and under high-liquidity swap conditions the users are initialized with 18,000 units of each token. Each simulation is run over 3,000 epochs.

The Q-Learning algorithm requires discrete values of the observation. Therefore the continuous - real state value is discretized by placing it into 1 of 500 buckets over slippage values ranging from -20\% to 20\% (negative slippage indicates more tokens received than spent). The environment is reset to a random state, along with random fee rate \& leverage coefficient values at the beginning of each epoch.

\subsection{Generation of Users} \label{userGen}
Users are generated as a list of objects, $U$, from the User class. Each user is assigned a balance of 1,000 units of each token when the environment is reset. The users are assigned an identification number - equal to their 0-indexed position in the list. There are two additional member variables that a user object has, namely, slippage tolerance and urgency. 
\subsubsection{Slippage Tolerance}
There are two different modes of slippage tolerance generation for the user objects.
\begin{itemize}
    \item Normal = Here, the slippage tolerances are assigned by random sampling over a truncated normal distribution with $mean(\mu) = 0.25$, $standard\ deviation(\sigma) = 0.25$, $lower\ bound = 0.1$, and $upper\ bound = 5$.
    \item Loose = Here, the slippage tolerances are assigned by random sampling over a truncated normal distribution with $mean(\mu) = 0.75$, $standard\ deviation(\sigma) = 0.75$, $lower\ bound = 0.1$, and $upper\ bound = 5$.
\end{itemize}

Only one of these two is used at any given time to run a simulation. This variation in the slippage tolerances allows us to study the sensitivity of different users to the changes in the simulated market conditions.

\subsubsection{Urgency}
Each user is also assigned an urgency factor, $\upsilon$, to modify their slippage tolerance, $\tau$. 
\begin{equation}
    \tau = \tau_{t-1} * e^{\upsilon}
\end{equation}
A user with a high $\upsilon$ indicates that they need to make the swap urgently, and they are less likely to wait for the market conditions to change in hopes of obtaining a better price. On the other hand, a user with a lower $\upsilon$ is more likely to wait out the market - till they get the slippage they are looking for. 

These urgency factors are assigned to the user objects by random sampling over a truncated normal distribution with $mean(\mu) = ln(1.5)$, $standard\ deviation(\sigma) = 0.25$, $lower\ bound = ln(1)$, and $upper\ bound = ln(2)$.

\begin{figure*}[!ht]
    \centering
    \setkeys{Gin}{width=\linewidth}
\begin{subfigure}{0.245\textwidth}
    \caption*{\bfseries Baseline}
\includegraphics{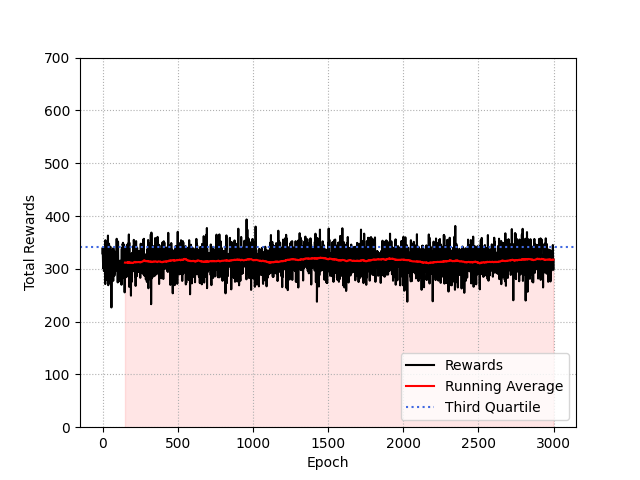}
\includegraphics{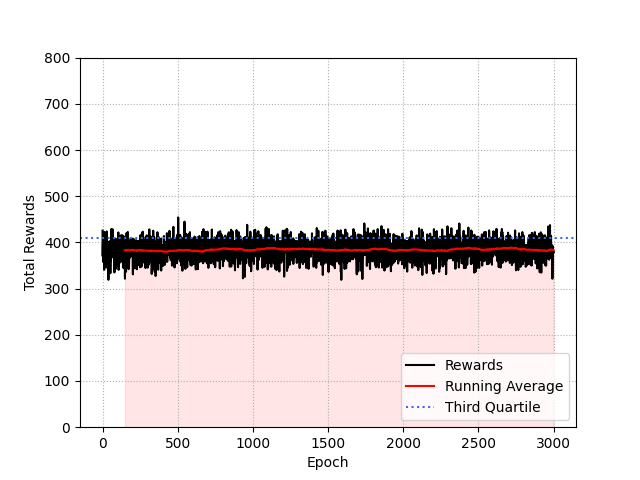}
\end{subfigure}
\hfil
\begin{subfigure}{0.245\linewidth}
    \caption*{\bfseries Agent 1}
\includegraphics{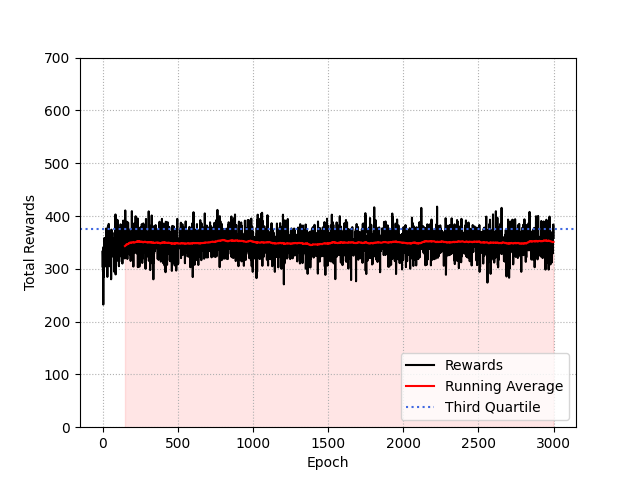}
\includegraphics{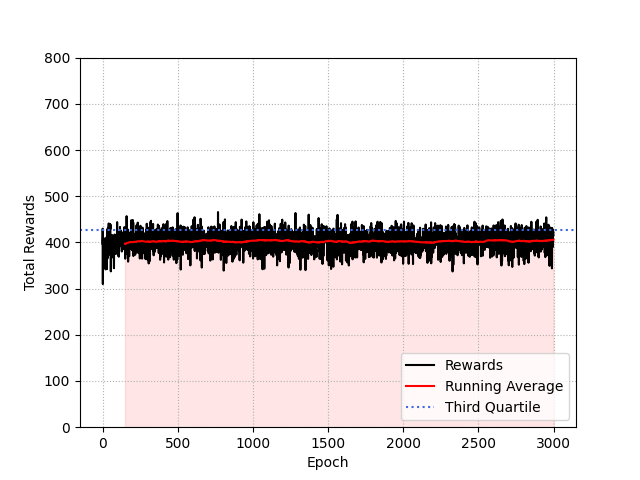}
\end{subfigure}
\hfil
\begin{subfigure}{0.245\linewidth}
    \caption*{\bfseries Agent 2}
\includegraphics{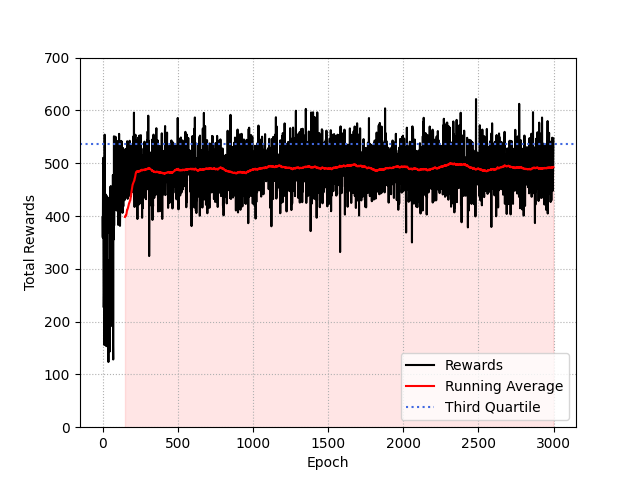}
\includegraphics{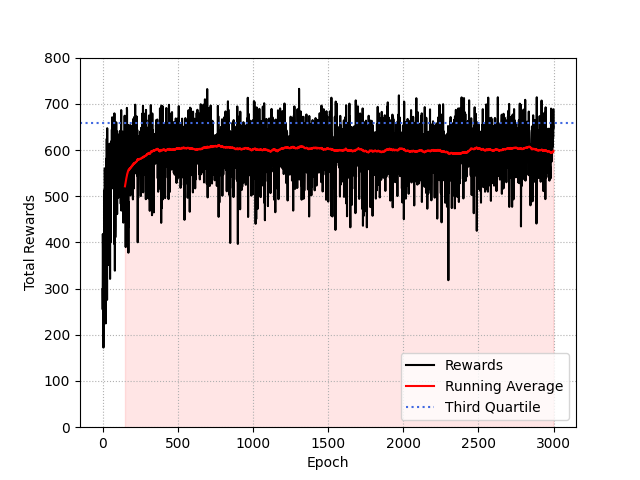}
\end{subfigure}
\hfil
\begin{subfigure}{0.245\linewidth}
    \caption*{\bfseries QLAMMP}
\includegraphics{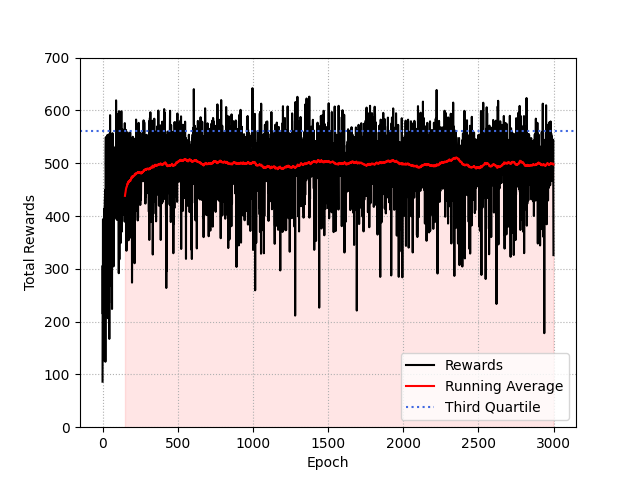}
\includegraphics{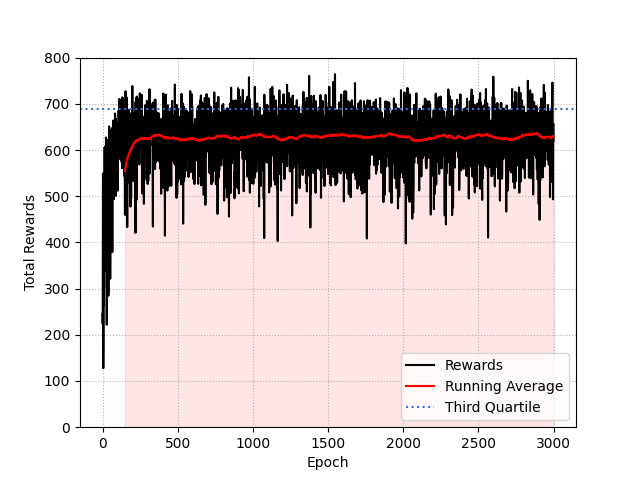}
\end{subfigure}
    \caption{Row 1 shows the agents' performance in a system with normal slippage tolerances. Row 2 shows the agents' performance in a system with loose slippage tolerances. QLAMMP can be seen significantly outperforming its counterparts in both environments.}
    \label{fig:comparison}
    \end{figure*}

\subsection{Swaps}
\subsubsection{Generation}
Swaps are generated by the environment at the beginning of each epoch, when the \texttt{setCurve}() method is called. Each swap is an object of the Swap class, and each object has the following seven member variables.
\begin{itemize}
    \item tolerance($\tau$) = this denotes the slippage tolerance that the user undertaking this swap will have. Once one of the aforementioned two modes has been chosen for the simulation, random samples are drawn from the chosen distribution and allotted to each swap. 
    \item urgency($\upsilon$) = this denotes the urgency factor that dictates the amount of leeway a user will allow for in their assigned tolerance.
    \item userNum = this value is used to assign the swap to a user in the system. It is the same as the user's index in the 0-indexed user list.
    
    The following member variables are only used if the user ``\textit{holds}" the transaction and is pushed back in the swap queue.
    \item amt = this denotes the amount that the user chose in the first iteration of querying the protocol for the given swap.
    \item idx = this denotes the token index that the user chose in the first iteration of querying the protocol for the given swap.
    \item tries = this is a counter variable that tracks the number of queries made to the protocol with regard to the given swap.
    \item new = this is a flag variable that denotes whether the swap is one that was pushed back into the queue prior to the current query. If set to False, this is a returning swap, and all the above variables need to be used for the swap.
\end{itemize}

\subsubsection{Handling Swaps}
As described above, we now have swaps that have been randomly generated. Each swap is randomly assigned to a user from $U = \{U_1, U_2, ..., U_\mathcal{N}\}$, via a discrete uniform sampling and added to the swap queue, $\mathcal{E}$.

For our simulations, each epoch consists of 400 steps. The swap to be executed in a given step is decided by the swap object, $\mathcal{E}_i$, at the head of queue $\mathcal{E}$. The user specified by $\mathcal{E}_i$ attempts to make the swap - if successful, the user returns status code 1; if holding, the user returns status code 0; and if the swap fails, the user returns status code -1. $\mathcal{E}_i$ is then popped out of queue $\mathcal{E}$.

\subsection{Metrics}
\subsubsection{Evaluation Metric}
We model the agent to optimize for fees. The evaluation of the agent's performance toward this goal can be done by looking at two key factors, (1) the total fee accrued over an epoch and  (2) the number of swaps canceled in an epoch. Our reward function incorporates both these factors, and we look at the sum of all rewards awarded to an agent over an epoch. We can evaluate each agent based on the change in this value.

\subsubsection{Baseline}
We use the same environment to generate our baseline performance. For this purpose, there is no agent, no actions are taken, and the environment operates statically with a fee rate of 0.17\% and a leverage coefficient of 42. Both these values represent the arithmetic means of their respective domains. The rest of the environment functions exactly the same as it would otherwise.

\subsection{Simulation}
We simulate the three agents individually and observe their performance across the different slippage tolerance modes against their respective baseline performances. We expect the agents to adapt to the user's demands in a way that maximizes profit by extracting the maximum fee possible while ensuring that canceled transactions are kept at a minimum.

We hypothesize that our agents will excel at different components of the given optimization problem. Agent 1, who controls the rate of fees, is hypothesized to perform better when the fee rate primarily governs the slippage. This case is likely to occur when the amounts being swapped are a small fraction of the total liquidity in the pool - the invariant mostly stays close to the ``center" of the curve, where the curve is `flatter.' Hence the slippage is dictated by the fees being levied on the swap. Agent 2, on the other hand, will be more effective at optimizing fees when users make swaps of amounts that represent more significant fractions of the total liquidity in the system - the invariant is pushed closer to the axes, where the curve starts deviating from its constant sum characteristic. This means the behavior of the curve primarily dictates the slippage, and Agent 2 can optimize for this by adjusting the curve's leverage coefficient. QLAMMP makes actions that change both the fee rate and the leverage coefficient. Thus it is expected to combine the expertise of Agents 1 \& 2 and outperform both across the given tasks.

We also expect our proposed agent, QLAMMP, to be able to adapt to time-varied user behavior. To accomplish this, one may modify the users' slippage tolerance from \textit{loose} to \textit{normal} and the other way around -- halfway through the epochs. This shift would mean that the agent would adapt to optimize for a particular type of user but would face a completely different user after the halfway mark. We hypothesize that an effective agent would adapt to this change quickly and extract maximum fee by changing its policy.

\section{Results}

The performance of the proposed agent, QLAMMP, is assessed with respect to the two modes of slippage tolerance, a high liquidity swap condition and a varying user demand condition.

\subsection{Varying Slippage Tolerance}
We assess the agents' performance against a system of users with normal and loose slippage tolerances, respectively. Row~1 of Fig.~\ref{fig:comparison} shows the agents' behavior against users with a normal slippage tolerance. Row~2 of Fig.~\ref{fig:comparison} shows the agents' behavior against users with a loose slippage tolerance. Both conditions corroborate the hypothesis that Agent 1 cannot make a significant difference in total rewards because the price impact, $\varphi$, is primarily governed by the fee rate, $\mathcal{R}_f$. 

\begin{figure}[ht]
\begin{subfigure}{.24\textwidth}
  \centering
  \includegraphics[width=1.04\linewidth]{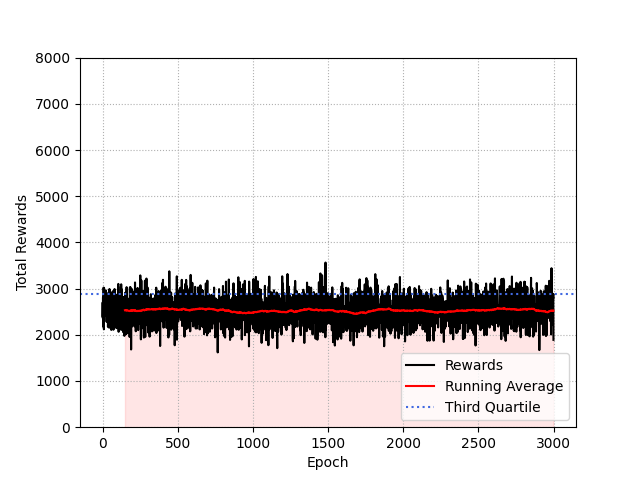}  
  \caption*{\bfseries Baseline}
  \label{fig:sub-first1}
\end{subfigure}
\hfill
\begin{subfigure}{.24\textwidth}
  \centering
  \includegraphics[width=1.04\linewidth]{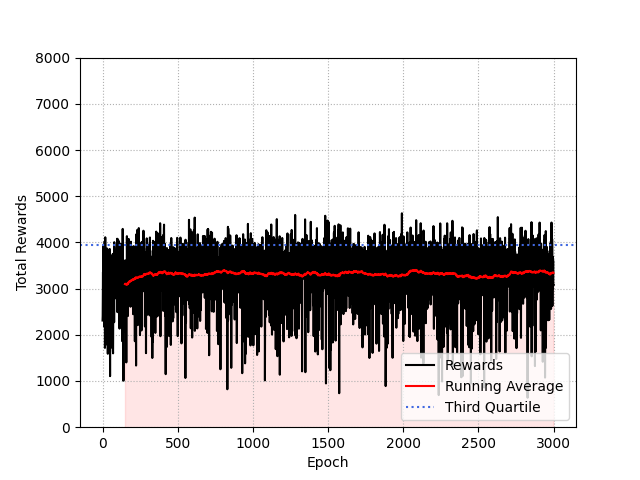}  
  \caption*{\bfseries Agent 1}
  \label{fig:sub-second2}
\end{subfigure}
\begin{subfigure}{.24\textwidth}
  \centering
  \includegraphics[width=1.04\linewidth]{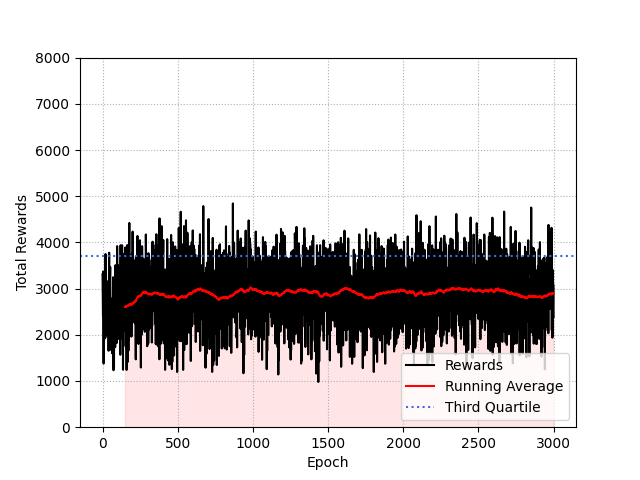}  
  \caption*{\bfseries Agent 2}
  \label{fig:sub-third3}
\end{subfigure}
\hfill
\begin{subfigure}{.24\textwidth}
  \centering
  \includegraphics[width=1.04\linewidth]{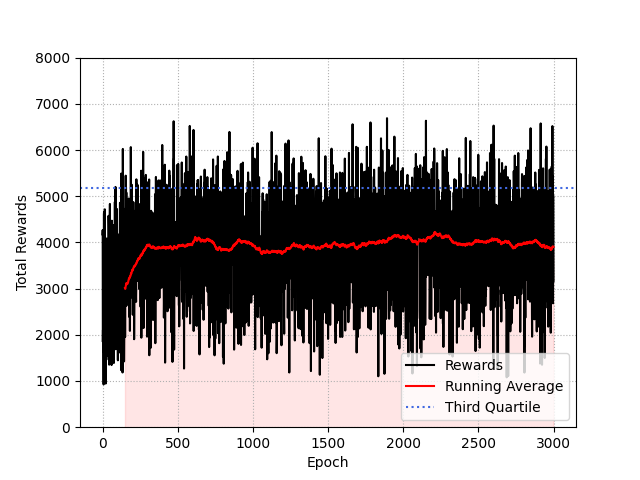}  
  \caption*{\bfseries QLAMMP}
  \label{fig:sub-fourth4}
\end{subfigure}
\caption{Agent performance in high-liquidity swap environments. QLAMMP is observed to significantly outperform the other agents when faced with high-liquidity swap demands.}
\label{fig:high}
\end{figure}

The hypothesis is further concretized by Agent 2's performance, which is significantly better than the baseline and Agent 1. This boost in rewards can directly be attributed to the control Agent 2 has over the fee rate, $\mathcal{R}_f$. QLAMMP performs significantly better than the baseline, and we can ascribe its success to the $\mathcal{R}_f$-modifying subset of its action set. The slight improvement of QLAMMP over Agent 2 is likely due to its additional control over the leverage coefficient, $\mathcal{A}$.

\subsection{High Liquidity Swap}
We also assess the agents' performance in conditions where each user swaps a relatively large fraction of tokens compared to the liquidity in the pool. Fig.~\ref{fig:high} shows the agents' performance when these users comprise the environment. Compared to the earlier simulation, the users now make trades of up to \$18,000 in initial pools of \$20,000. As hypothesized, Agent 1 outperforms the baseline and Agent 2 because as the pool state moves to a point closer to the axes, the price impact, $\varphi$, increases and is governed primarily by the leverage coefficient, $\mathcal{A}$. Agent 2 lacks control over $\mathcal{A}$ and thus cannot adapt sufficiently to the users' demands. Once again, QLAMMP significantly outperforms the baseline, Agent 1, and Agent 2 due to its larger action space - controlling both $\mathcal{A}$ and $\mathcal{R}_f$.

\begin{figure}[ht]
\begin{subfigure}{.24\textwidth}
\caption*{\bfseries Baseline}
  \centering
  \includegraphics[width=1.04\linewidth]{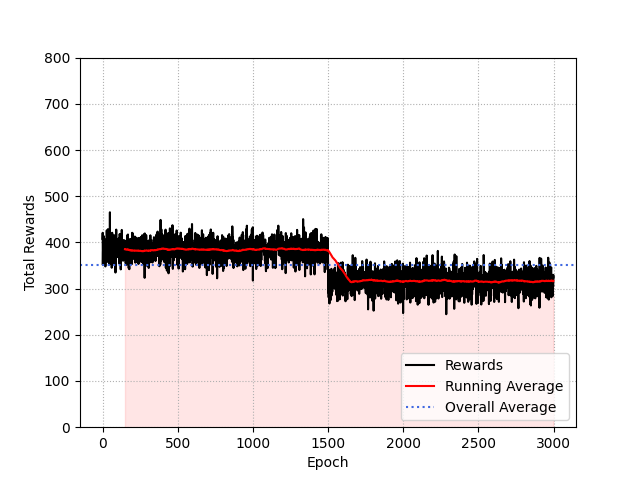}  
  \label{fig:sub-first}
\end{subfigure}
\hfill
\begin{subfigure}{.24\textwidth}
\caption*{\bfseries QLAMMP}
  \centering
  \includegraphics[width=1.04\linewidth]{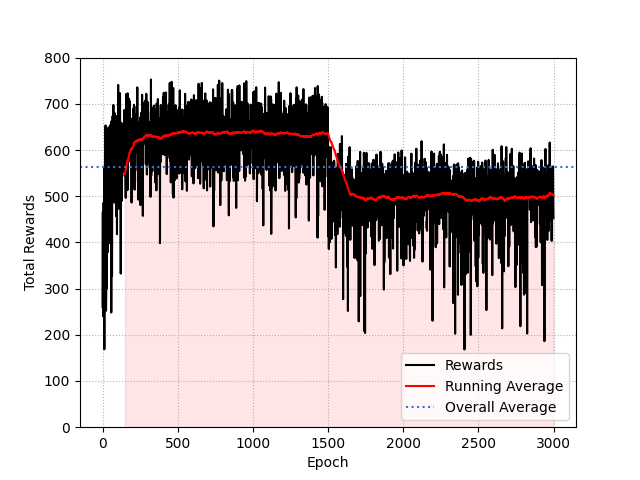}  
  \label{fig:sub-second}
\end{subfigure}
\begin{subfigure}{.24\textwidth}
  \centering
  \includegraphics[width=1.04\linewidth]{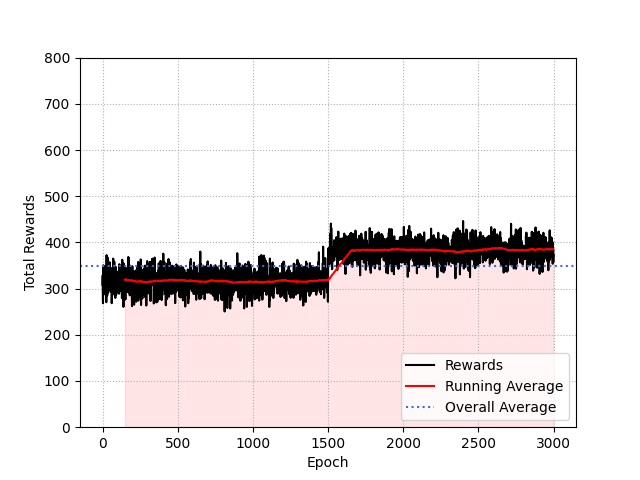}  
  \label{fig:sub-third}
\end{subfigure}
\hfill
\begin{subfigure}{.24\textwidth}
  \centering
  \includegraphics[width=1.04\linewidth]{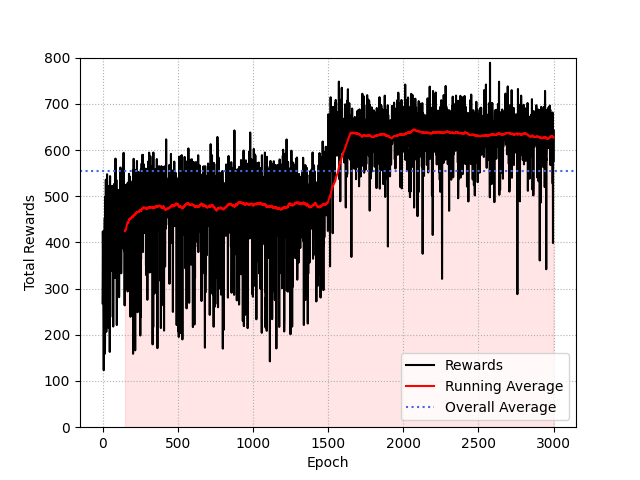}  
  \label{fig:sub-fourth}
\end{subfigure}
\caption{Row 1 shows the user behavior changing from loose to normal slippage tolerances. Row 2 shows the user behavior changing from normal to loose slippage tolerances.}
\label{fig:change}
\end{figure}

\subsection{User Behavior Change}
QLAMMP also performs well when user behavior changes from a normal slippage tolerance to a loose slippage tolerance and vice-versa. We observe that QLAMMP can quickly adapt to the new market conditions and update the protocol parameters to extract fees from the users optimally. Fig.~\ref{fig:change} shows QLAMMP adapting its policy to suit the new market conditions.

\begin{figure}[ht]
  \begin{subfigure}{.24\textwidth}
  \centering
  \includegraphics[width=1.04\linewidth]{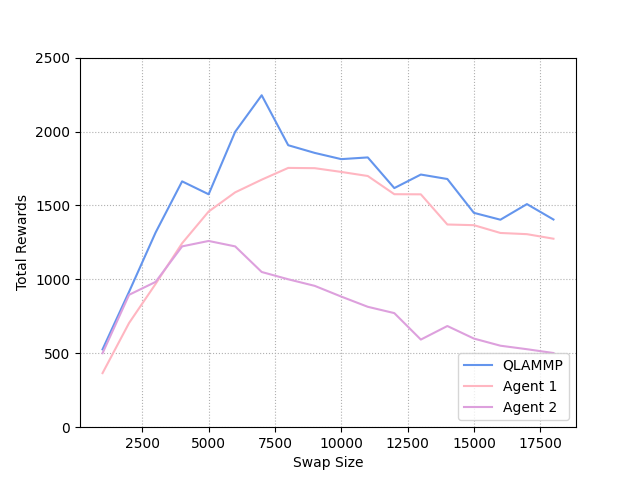}
  \centering
  \caption{Rewards v. Swap Size}
  \label{fig:sub-1}
\end{subfigure}
\hfill
\begin{subfigure}{.24\textwidth}
  \centering
  \includegraphics[width=1.04\linewidth]{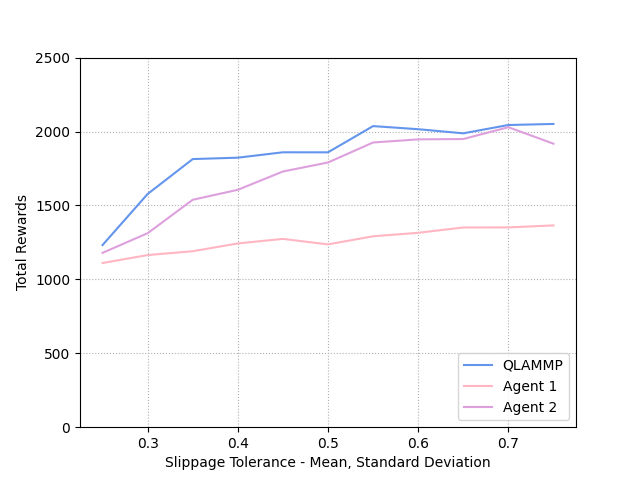}  
  \caption{Rewards v. Slippage Tolerance}
  \label{fig:sub-2}
\end{subfigure}
  \begin{subfigure}{.24\textwidth}
  \centering
  \includegraphics[width=1.04\linewidth]{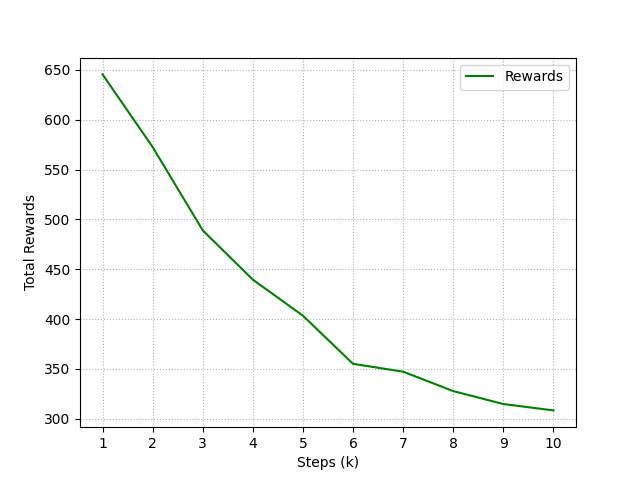}
  \centering
  \caption{Rewards v. update interval}
  \label{fig:sub-first5}
\end{subfigure}
\hfill
\begin{subfigure}{.24\textwidth}
  \centering
  \includegraphics[width=1.04\linewidth]{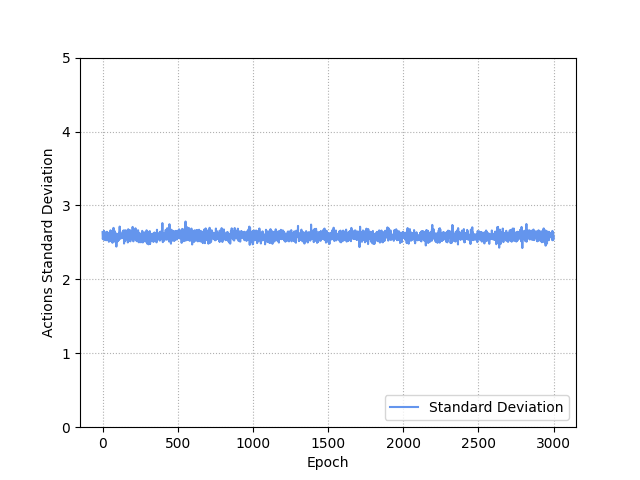}  
  \caption{Standard deviation of actions}
  \label{fig:sub-second6}
\end{subfigure}
\caption{Other Findings}
\label{fig:update}
\end{figure}

\subsection{Other Findings}
\subsubsection{Swap Size}
We simulate the agents using a range of swap amounts between \$1,000 and \$18,000, both inclusive. Our hypothesis is based on the assumption that the agents' performance varies across swap sizes. This hypothesis is confirmed by Fig.~\ref{fig:sub-1}. We see that for smaller amounts, Agent 2 has superior performance, and beyond \$3,750, Agent 1 takes over and begins outperforming Agent 2. As expected, QLAMMP consistently outperforms both Agents 1 and 2.

\subsubsection{Slippage Tolerance}
We also simulate the agents using a range of slippage tolerances samplings. These samplings are made with means and standard deviations ranging from 0.25 to 0.75, both inclusive. Our hypothesis - that the users' tolerance affects the rewards accrued by the agents - is confirmed by Fig.~\ref{fig:sub-2}. We can see that all three agents can adapt the protocol to earn more rewards as the users' slippage tolerance increases.   

\subsubsection{Update Intervals}
A practical consideration for deploying such an agent on-chain requires a decision on how often the parameters will be updated. We look at the total rewards collected by the agent as a function of the update interval. As expected, an almost-linear relation is observed. The developer of such a protocol must consider a trade-off between the update interval and the agent's efficacy. Fig.~\ref{fig:sub-first5} shows QLAMMP's performance as a function of the update interval, $k$.

\subsubsection{Policy Evaluation}
Since the agent makes an exploitative choice at each decision epoch, $t$, we look at the mean standard deviation of the agent's actions across the epochs. Fig.~\ref{fig:sub-second6} shows that QLAMMP's mean standard deviation stays close to 2.64. This value indicates that the agent is adapting the protocol's behavior to suit the current market demands rather than just sticking to a given protocol configuration for which it expects the highest reward over all the epochs.

\section{Discussion \& Future Work}
In this paper, we present QLAMMP, a reinforcement learning agent for optimizing fees on Automated Market Making Protocols. We empirically show that the proposed Q-learning agents significantly outperform their static counterparts in fee collection. Furthermore, we also show that the agents can adapt to changing user mechanics and are robust under various scenarios. To achieve this gain in fee collection, the smart contract must frequently update the protocol parameters. This computation will likely incur gas fees; therefore, we also looked at updating protocol parameters every $k$ steps, thereby reducing computation. As expected, the agents' performance drops as $k$ increases - implying a direct relation between computation and performance. However, the agents still consistently outperformed their baselines in all cases. The agent's computation could also be offloaded and accessed using an oracle - this consideration is left to the developer based on practical considerations. Our open source implementation proves the concept's soundness and demonstrates the robustness of the agent under varied conditions.

On-chain deployment of QLAMMP can primarily be done in two ways. 
\begin{itemize}
    \item Deploying a Smart Contract with the Q-table stored on it. This method would allow the protocol to automatically update the parameters after each trade. However, such an offline Q-table would need to be updated with new values at frequent intervals to ensure it is up to date with the current market movements.
    \item To solve the above problem, one could deploy an online agent off-chain and use an oracle to update the protocol parameters at given intervals. This oracle should ideally be decentralized to preserve the core principles of the DEX~\cite{ma2019reliable, breidenbach2021chainlink}. Perhaps such an oracle would use its own consensus mechanism, choosing the median value of all the proposed values to update the protocol.
\end{itemize}

In the future, we plan on using deep reinforcement learning to train the agent. We also plan on optimizing the agents' computations and looking at efficient methods of on-chain deployment of the proposed agent, which are the keys to such agents' practical usability.

\bibliography{ref}
\bibliographystyle{ieeetr}

\end{document}